\documentclass[letterpaper, 10 pt, conference]{ieeeconf}  

\IEEEoverridecommandlockouts                              

\let\labelindent\relax
\usepackage{graphicx}
\usepackage{wrapfig}
\usepackage{picinpar}
\usepackage{float}
\usepackage{algorithmic}
\usepackage{algorithm}
\usepackage{longtable}
\usepackage{verbatim}
\usepackage{multirow}
\usepackage{graphicx}
\usepackage{booktabs}
\usepackage{diagbox}
\usepackage{bm}
\usepackage{amsfonts,amssymb}
\usepackage{makecell}
\usepackage{cite}
\usepackage{amsmath}
\usepackage{enumerate}
\usepackage{color}
\usepackage{soul}
\usepackage{verbatim}
\usepackage{enumitem}
\usepackage{siunitx}
\usepackage{threeparttable}
\usepackage{graphicx}
\usepackage{makecell}
\usepackage{amsmath}
\usepackage{mathrsfs}
\usepackage{hyperref}
\hypersetup{colorlinks = true,
	linkcolor = blue,
	urlcolor = blue,
	citecolor = blue} 
\usepackage{array} 

\newcolumntype{M}[1]{>{\centering\arraybackslash}p{#1}}

\title{\LARGE \bf
Pretraining-finetuning Framework for Efficient Co-design: A Case Study on Quadruped Robot Parkour
}

\author{Ci Chen$^1$, Jiyu Yu$^1$, Chao Li$^2$, Haojian Lu$^1$, Hongbo Gao$^3$, Rong Xiong$^1$, Yue Wang$^1$
\thanks{*This work was supported in part by the National Natural Science Foundation of China under Grant U2013601. Corresponding author: Yue Wang ({\tt\small wangyue@iipc.zju.edu.cn})}
\thanks{$^{1}$Ci Chen, Jiyu Yu, Haojian Lu, Rong Xiong, and Yue Wang are with the State Key Laboratory of Industrial Control and Technology, Zhejiang University, Hangzhou, 310027, China.}
\thanks{$^{2}$Chao Li is with the DeepRobotics Company, Hangzhou, 310058, China.}
\thanks{$^{3}$Hongbo Gao is with the School of Information Science and Technology, University of Science and Technology of China, Hefei, 230026, China.}
}

\begin{document}

\maketitle
\thispagestyle{empty}
\pagestyle{empty}

\begin{abstract}
In nature, animals with exceptional locomotion abilities, such as cougars, often possess asymmetric fore and hind legs. This observation inspired us: could optimizing the leg length of quadruped robots endow them with similar locomotive capabilities? In this paper, we propose an approach that co-optimizes the mechanical structure and control policy to boost the locomotive prowess of quadruped robots. Specifically, we introduce a novel pretraining-finetuning framework, which not only guarantees optimal control strategies for each mechanical candidate but also ensures time efficiency. Additionally, we have devised an innovative training method for our pretraining network, integrating spatial domain randomization with regularization methods, markedly improving the network's generalizability. 
Our experimental results indicate that the proposed pretraining-finetuning framework significantly enhances the overall co-design performance with less time consumption. Moreover, the co-design strategy substantially exceeds the conventional method of independently optimizing control strategies, further improving the robot's locomotive performance and providing an innovative approach to enhancing the extreme parkour capabilities of quadruped robots.
\end{abstract}

\section{INTRODUCTION}

Quadruped animals exhibit exceptional mobility capabilities. 
For instance, a red kangaroo, about 1.8 meters tall, can easily leap forward more than 9 meters, quintupling its height. It is noteworthy that quadruped mammals typically exhibit an asymmetric leg structure, a result of natural evolution that favors agile movement in their natural habitats \cite{gupta2021embodied}.
This realization spurred our curiosity: Could adjust the leg length of standard quadruped robots potentially unlock greater locomotive capabilities?

\begin{figure}[t]
	\centering
	\includegraphics[width=0.45\textwidth]{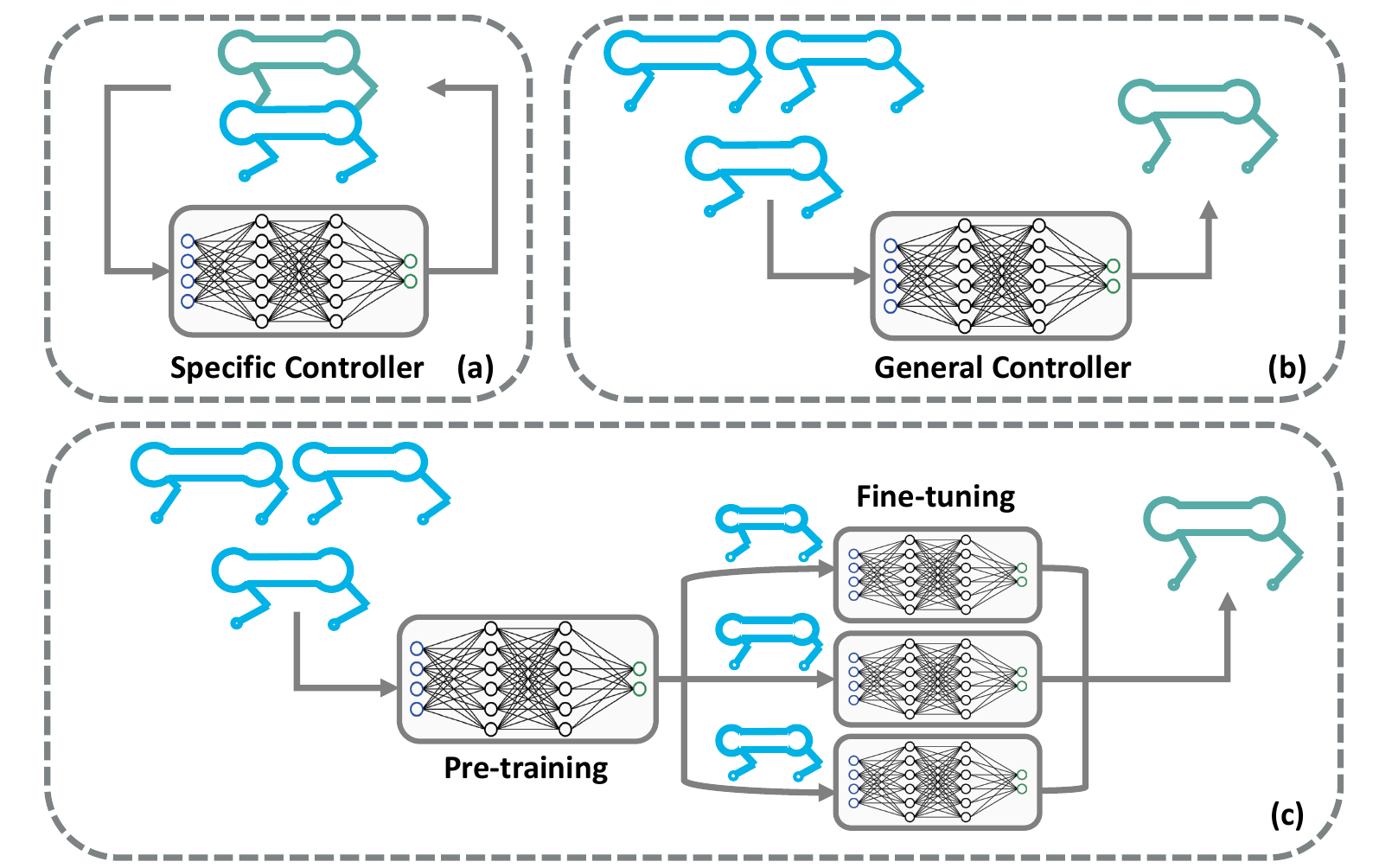}
	\setlength{\abovecaptionskip}{-0.2em}
	\caption{Comparison of co-design methods. (a) Designing distinct control strategies tailored to various morphologies. (b) Utilizing a generalized strategy as a substitute. (c) The algorithm we introduce starts by training a generalized model across varying leg lengths, followed by fine-tuning for specific morphology. }
	\vspace{-2.0em}
	\label{compare}
\end{figure}

The joint optimization of a robot's mechanical structure and control policy is referred to as \emph{co-design}, and currently, there are numerous works focused on the co-design of quadruped robots \cite{belmonte2022meta,chen2023c,chen2023deep,yu2023multi,bjelonic2023learning}. These studies are primarily confined to walking tasks. For instance, \cite{belmonte2022meta} explored the ideal proportion between thighs and shanks, optimizing for objectives such as speed tracking and energy efficacy across varied terrains. \cite{yu2023multi} accentuated the generalizability of the transformer-based control strategy, asserting its adaptability across a myriad of leg length ratios. It also showcased that using a fixed control policy, solely morphological updates could enable a robot to step over a 10 \si{cm} platform. \cite{chen2023c,chen2023deep} improved the mobility of small-scale quadruped robots with limited degrees of freedom on flat ground through co-design approaches. However, the potential of robot co-design has not yet been fully realized in simple walking tasks. In more challenging control tasks like parkour tasks \cite{zhuang2023robot,hoeller2023anymal,cheng2023extreme,rudin2022advanced}, 
such as high jumping and long jumping, the significance of co-design becomes even more prominent. The physical capabilities of a robot and the scope of tasks it can execute are constrained by its hardware components. When the mechanical structure is not aptly designed, even the most ingeniously crafted motion control policies might struggle to fulfill the intended tasks. It is through a reciprocal enhancement of mechanical structure and control policies that the functional ceiling of a robot can be significantly elevated.

The co-design of robots can be primarily divided into methods based on dynamic models \cite{ha2017joint,geilinger2018skaterbots,fadini2021computational} and methods based on deep reinforcement learning (DRL) \cite{belmonte2022meta,chen2023c,chen2023deep,yu2023multi,bjelonic2023learning,hejna2020task,wang2018neural,gupta2021embodied,luck2020data,schaff2019jointly}. However, dynamic model-based methods require a wealth of expert knowledge and engineering experience to construct dynamical models and design equations and inequality constraints. For robots with different mechanical structures, system identification \cite{nagarajan2009state} must also be incorporated for motion control, which can be exceedingly tedious and complicated. 
DRL-based methods typically frame the problem as a bi-level optimization issue, with the lower-level optimizing the control strategy for a specified morphology and the upper-level searching for the best morphology given the optimal control policy. Depending on how the control strategy is obtained, there are mainly two categories. 
The first involves training specific controllers for given candidate structures during the upper-level optimization, as shown in Fig. \ref{compare}(a). For instance, \cite{gupta2021embodied} uses a tournament algorithm at the upper-level, employing the fitness of each candidate morphology as input to derive the optimal morphology. On the lower-level, it utilizes 1152 CPUs with DRL algorithms to learn the optimal control policies for various candidate morphologies, which is highly time-consuming and requires substantial computational power. 
In contrast, the second category focuses on the development of a versatile and generalizable model \cite{chen2023c,belmonte2022meta,yu2023multi,bjelonic2023learning}, which is capable of providing alternative strategies applicable to various morphologies, as illustrated in Fig. \ref{compare}(b). Specifically, \cite{chen2023c} employs a concurrent network architecture, \cite{bjelonic2023learning} adopts a teacher-student architecture, and \cite{yu2023multi} leverages a transformer architecture. However, this approach of substituting optimal strategies with generalized ones, while time-saving, cannot ensure optimality in specific morphologies, potentially impairing the efficacy of upper-level optimization decisions.

\begin{figure}[t]
	\centering
	\includegraphics[width=0.48\textwidth]{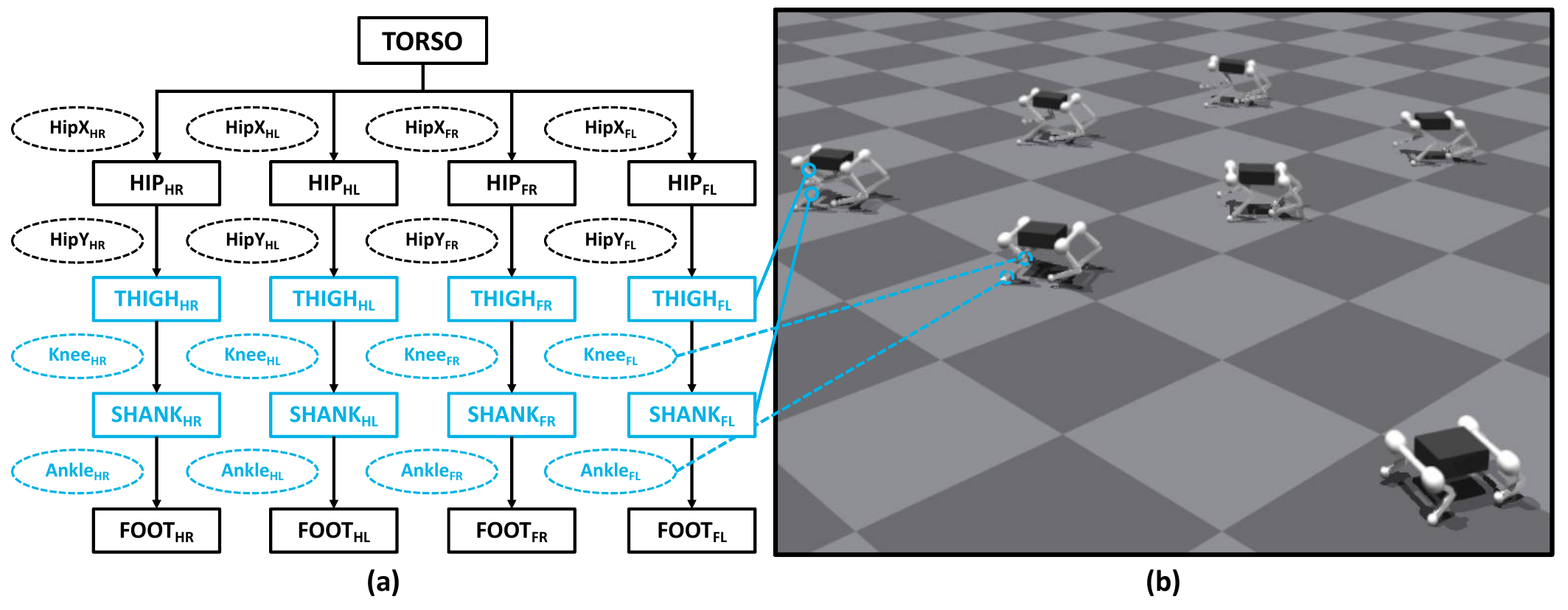}
	\setlength{\abovecaptionskip}{-1.5em}
	\caption{(a) The URDF model of the quadruped robot, where the solid rectangles represent links, and the dashed ellipses represent joints. The black parts denote fixed parameters, while the blue parts denote parameters that need to be changed. (b) The quadruped robot with different scaling parameters.}
	\vspace{-2.0em}
	\label{urdf}
\end{figure}

In order to maintain optimality while also conserving computational power, in this paper, we introduced a pretraining-finetuning framework for quadruped robots’ co-design, with the goal of improving their parkour performance, as shown in Fig. \ref{compare}(c). 
We first trained a parkour strategy that is adaptable across a variety of morphologies. Specifically, we integrated spatial domain randomization and regularization techniques. The former enables simultaneous training of thousands of robots with varying morphologies, while the latter helps in reducing memory biases in the value network. The combination of them significantly enhances the generalizability of the pre-trained model. 
In the subsequent phase of morphology optimization, we fine-tuned this generalized model to suit each specific morphology candidate. Our experimental results indicate that merely 400 steps of fine-tuning (6.67\% of the typical training steps) were sufficient for the new model to converge, delivering performance comparable to models specifically designed for each structure. In summary, the contributions of this paper are as follows:

\begin{enumerate}
	\item{We introduced a pretraining-finetuning framework to address the co-design challenge in robotics, ensuring optimality while conserving computational resources.}
	\item{We proposed the concept of combining spatial domain randomization with regularization to obtain a pre-trained model, effectively enabling generalization across robots with different leg lengths.}
	\item{Extensive ablation and comparative experiments have validated the efficacy of our proposed algorithm in co-design tasks, demonstrating its capability to improve the parkour performance of robots.}
\end{enumerate}

\section{Methodology}
\begin{figure*}[t]
	\centering
	\includegraphics[width=1.0\textwidth]{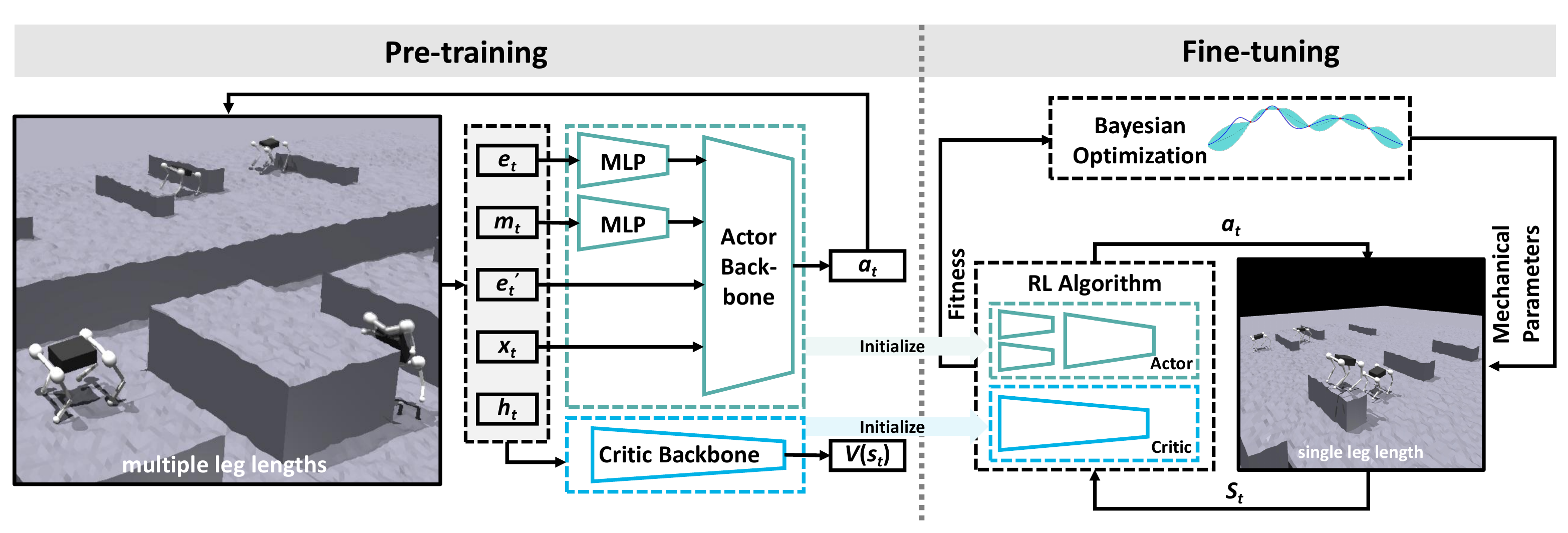}
	\setlength{\abovecaptionskip}{-2em}
	\caption{
		Overview of the proposed co-design pipeline. We initially pre-train a control strategy capable of generalizing across various morphologies. Subsequently, we embed the fine-tuning process into the morphological optimization phase, fine-tuning the pre-trained strategy for each set of candidate morphologies. This approach provides more accurate fitness values for morphological optimization.
		We employed the Proximal Policy Optimization (PPO) algorithm to update our policy, utilizing an asymmetric Actor-Critic architecture.
	}
	\vspace{-1em}
	\label{framework}
\end{figure*}

\subsection{System Overview}
The overall framework of our method is depicted in Fig. \ref{framework}, including two phases: Pre-training (Section. \ref{pre-training}) and Bayesian optimization (BO) with embedded fine-tuning (Section. \ref{BO}). 
We frame the parkour control problem as a Markov Decision Process (MDP), which is defined by a tuple $\left\langle {\mathcal S, \mathcal A, P, r, \gamma } \right\rangle $,
in which $\mathcal S$ is the state space, $\mathcal A$ is the action space, $P(s_{t+1}|s_{t},a_{t})$ is the transition probability, $r(s_{t},a_{t})$ is the reward function, and $\gamma$ is the discount factor.
At each timestep, the agent receives a state  $s_{t} \in \mathcal{S}$ from the environment and selects an action $a_{t} \in \mathcal{A}$, guided by policy $\pi_{\theta}(a_{t}|s_{t})$. This action yields a reward $r_t$. The objective of DRL is to derive an optimal policy $\pi^{*}$ that maximizes the expected sum of discounted rewards ${\mathbb{E}}\left[ {\sum_{t=0}^{T}  {{\gamma ^t}{r_t}} } \right]$.

\textbf{Design of State and Action Space.}
The state comprises four parts: the proprioceptive state $x_t$, which includes the body’s angular velocity, roll angle, pitch angle, the difference between the body’s yaw angle and the target yaw angle, positions and velocities of joints, the actions in the previous step, and foot contact information; 
the explicit privileged state ${e'_t}$, which refers to the body’s linear velocity; 
the privileged state $e_t$, which encompasses the robot’s mass, center of mass position, robot morphology parameters, friction factor, and motor strength parameters; 
the exteroceptive state $m_t$, which consists of height sampling points; 
and the historical proprioceptive state $h_t$. 
The action $a_t$ is the target angles for the robot’s twelve joints, which are converted into torques through a PD controller and then transmitted to the robot.

\textbf{Design of Reward Function.}
Unlike previous works \cite{rudin2022learning,kumar2021rma,ji2022concurrent,nahrendra2023dreamwaq}, which utilized velocity tracking as the primary reward function item, manually specifying the robot’s linear velocity and its yaw angle in parkour tasks constraints the solution space of actions, leading to failure in performing parkour tasks. We designed our reward functions by referring to \cite{cheng2023extreme}. The primary reward function \emph{Goal Tracking}, encourages the robot to move toward waypoint markers. When encountering an elevated platform, this reward function item motivates the robot to jump onto it rather than bypass it, as in obstacle avoidance tasks. 
The design of the reward function item is shown in Section. \ref{reward_function} of the Appendix.

\subsection{Pre-training Phase}
\label{pre-training}

\begin{table}[t]
	\centering
	\caption{The parameters that need to be modified in the URDF file}
	\vspace{-1em}
	\begin{threeparttable}
		\resizebox{0.48\textwidth}{!}{
			\begin{tabular}{|c|c|c|c|}
					\hline
				Tag   & Attribute & Parameters & Values  \\
				\hline
				\multirow{9}[14]{*}{Link} & \multirow{5}[6]{*}{Inertial} & Origin & x:0, y: 0, z: $\frac{{ - {l_i} \times {\xi _i}}}{2}$ \\
				\cline{3-4}          &       & Mass  & value: ${m_i} \times {\xi _i}$ \\
				\cline{3-4}          &       & \multirow{3}[2]{*}{Inertia} & ixx: $\frac{{{m_i}}}{{12}}\left( {b_i^2 + l_i^2} \right)$ \\
				&       &       & iyy: $\frac{{{m_i}}}{{12}}\left( {b_i^2 + l_i^2} \right)$ \\
				&       &       & izz: $\frac{{{m_i}b_i^2}}{6}$ \\
				\cline{2-4}          & \multirow{2}[4]{*}{Visual} & Origin & x:0, y:0, z:$\frac{{ - {l_i} \times {\xi _i}}}{2}$ \\
				\cline{3-4}          &       & Geometry/Box & size: ${b_i},{b_i},{l_i} \times {\xi _i}$ \\
				\cline{2-4}          & \multirow{2}[4]{*}{collision} & Origin & x:0, y:0, z: $\frac{{ - {l_i} \times {\xi _i}}}{2}$ \\
				\cline{3-4}          &       & Geometry/Box & size: ${b_i},{b_i},{l_i} \times {\xi _i}$\\
				\hline
				Knee joint & /     & Origin & x:0, y:0, z: ${z_{knee}} \times {\xi _{i,i = 0,2}}$\\
				\hline
				Ankle joint & /     & Origin & x:0, y:0, z:${z_{ankle}} \times {\xi _{i,i = 1,3}}$ \\
				\hline
			\end{tabular}
		}
		\begin{tablenotes}
			\footnotesize
			\begin{tabular}{p{0.48\textwidth}} 
				\item when $i$=0, it represents the left/right front thigh; $i$=1 corresponds to the left/right front shank; $i$=2 denotes the left/right hind thigh; and $i$=3 indicates the left/right hind shank. The variable $l_i$ represents the original length of each leg, while $m_i$ signifies the original mass of each leg. The collision model of the leg is represented by a cuboid with a square cross-section, where $b_i$ denotes the side length of the square cross-section. Furthermore, $z_{knee}$ represents the original position of the knee joint, and $z_{ankle}$ denotes the original position of the ankle joint.
			\end{tabular}
		\end{tablenotes}  
	\end{threeparttable}
	\label{URDF}
	\vspace{-3em}
\end{table}

\subsubsection {Spatial Domain Randomization}
\label{Spatial-DR}
Domain randomization emerges as a pivotal technique for augmenting the generalization capacity of DRL models\cite{shi2022reinforcement,ji2022concurrent}. By sampling dynamic parameters within a specified reasonable range, this method significantly enhances the model’s robustness. In traditional practices, domain randomization has involved altering variables such as friction coefficients and motor strengths during training to bridge the sim2real gap \cite{tan2018sim}, a process we refer to as \emph{Temporal Domain Randomization}. 
However, in our task, the structural parameters of the robot are determined when the simulator is loaded. Any changes to these parameters necessitate reloading the simulator. Making such modifications in every training epoch would undoubtedly significantly slow down the algorithm's training speed.
To address this, we introduce the concept of \emph{Spatial Domain Randomization}. Leveraging the extensive parallel capabilities of Isaac Gym \cite{makoviychuk2021isaac}, we implement this innovative approach by simultaneously training thousands of robots with varied morphologies in a single environment. This approach allows us to develop a parkour control strategy with good generalization in just a few hours.

Similar to our previous work \cite{chen2023deep}, we create robots with varying morphological parameters by modifying the unified robot description format (URDF) files of the quadruped robots. The URDF of a quadruped robot is illustrated in Fig. \ref{urdf}(a).
Although our method is applicable to various structural parameters of the robot, we have chosen to focus solely on the thigh and shank lengths for a clearer analysis of the optimization results, as indicated by the blue parts in Fig. \ref{urdf}(a). 
For each link and joint requiring modification, the specific parameters to be changed are detailed in Tab. \ref{URDF}. To ensure the robot's stability, we make the corresponding parts of the robot's left and right legs symmetrical. Therefore, the adjustable parameters are the scaling factors for the front thighs, front shanks, hind thighs, and hind shanks. When training the generalizable model, we allow the scaling factors to randomly sample within a feasible range, ${\xi _i} \sim \mathcal U(c_{min},c_{max})$. The sampled parameters ${\xi _{i}}$ are then used to construct robots with varied leg lengths, as illustrated in Fig. \ref{urdf}(b).
It's worth noting that when the length of a robot's limbs changes, the corresponding PD parameters must also be adjusted. Following the approach outlined in \cite{luo2024moral}, we utilize a polynomial to calculate the corrected proportional factors as follows:
\begin{equation}
{\eta _i} = a \times \xi _i^3 + b \times \xi _i^2 + c \times {\xi _i} + d
\end{equation}

The constants $a$, $b$, and $c$ are manually specified. Since we modified the thigh and shank, which correspond to the \textit{HipY} and \textit{Knee} joints shown in Fig. \ref{urdf}(a), we apply the correction factor ${\eta_i}$ to the PD parameters of these eight joints.

\subsubsection{Regularizations}
\label{REG}
The proposed spatial domain randomization method enables the algorithm to train using data from robots with varying leg lengths. However, as introduced in \cite{moon2022rethinking,cobbe2021phasic}, a value network trained across multiple agents is more likely to memorize the training data, leading to poor generalization to unvisited states. This not only hinders training performance but also negatively impacts testing performance in new environments. To mitigate this issue, 
we employed a \textit{Discount Regularization} approach, which adjusts the discount factor to a smaller value, denoted as $\gamma_{\text{reg}}$, where $0 < \gamma_{\text{reg}} < \gamma < 1$. The modification of the discount factor influences the calculation of the advantage function and temporal difference in the PPO algorithm, thereby affecting the computation of the Actor and Critic loss functions. The new formulations for the advantage function and temporal difference are as follows:
\begin{equation}
A_t^{{\gamma _{{\rm{reg}}}}} = \Sigma _{l = 0}^\infty {\left( {{\gamma _{{\rm{reg}}}}\lambda } \right)^l}\delta _{t + l}^{{\gamma _{{\rm{reg}}}}}
\end{equation}
\begin{equation}
\delta _t^{{\gamma _{{\rm{reg}}}}} = {r_t} + {\gamma _{{\rm{reg}}}}{V_\phi }\left( {{s_{t + 1}}} \right) - {V_\phi }\left( {{s_t}} \right)
\end{equation}
where $\lambda$ is the hyperparameter used in Generalized Advantage Estimation (GAE)\cite{schulman2015high}. Theoretically, a lower discount factor decreases the reliance on future rewards, allowing the algorithm to concentrate more on short-term gains. This approach can enhance generalization by reducing variance. For the mathematical proof of this method, please refer to the Section. \ref{regularization} of the Appendix.

\subsection{BO with Embedded Fine-tuning}
\label{BO}
\subsubsection{Fine-tuning Based on Pre-trained Model}
By integrating spatial domain randomization with regularization, we have developed a control strategy applicable to robots with varying leg lengths. However, this strategy does not guarantee optimality in each morphology. To provide a more reliable fitness function for the morphology optimization algorithm, each morphological candidate is fine-tuned. 
Unlike the training of generalization models, we removed the spatial domain randomization method during the fine-tuning process. Instead, we created thousands of identical robots to interact with the environment, allowing us to fine-tune specifically for the targeted morphology.

\subsubsection{Morphology Optimization}
The objective of morphology optimization is to find optimal morphology parameters $\xi \in \Xi $ for a specified parkour task $\kappa \in  \mathcal K $ and a given control strategy $\pi \in \Pi $, such that the objective function $f(\left. \xi \right|\kappa,\pi ):\Xi \times \mathcal K \times \Pi \to \mathbb R$ is maximized. This study focuses primarily on the high jump and long jump tasks, denoted as $\mathcal K = \left\{ {{\kappa_{high}},{\kappa_{long}}} \right\}$. The optimization objective (or \emph{fitness}) $f( \cdot )$ can be any metric, we use the robot’s non-discounted cumulative reward in a single epoch as our optimization objective, as shown below:
\begin{equation}
f = \frac{1}{N}\sum\nolimits_{i = 0}^N {\sum\nolimits_{t = 0}^T {r\left( {s_t^i,a_t^i} \right)} } 
\end{equation}
where $T$ denotes the number of steps in an episode, while $N$ represents the number of robots interacting with the environment. 
In our previous work \cite{chen2023c}, we demonstrated that the BO method is well-suited for optimizing morphological parameters in co-design, given that $\xi$ is non-differentiable with respect to $f(\cdot)$. We can summarize the optimization problem as follows:
\begin{equation}
{\xi ^*} = \mathop {\arg \max }\limits_{\xi \in \Xi } f(\left. \xi \right|\kappa,\pi _\xi)
\end{equation}

The BO algorithm is an iterative process where, in each iteration, a candidate morphology $\xi$ is selected. Based on this chosen morphology, the RL algorithm initially fine-tuned the control policy on the basis of the pre-trained model. After deriving the strategy $\pi _\xi$, the robot equipped with the morphology parameters $\xi$ interacts with the environment under the guidance of $\pi _\xi$ to obtain $f( \cdot )$. This outcome is then fed back to the BO algorithm, which continues to iterate until a specified number of rounds are completed, ultimately yielding the optimal morphology.

\section{Experiments}

\subsection{Implementation Details}
During the training process, we employed Isaac Gym as the simulator.
The network was built using PyTorch, and training was facilitated using Nvidia 3090 GPUs. The neural network policy generated PD targets for all motors, operating at a frequency of 50 \si{Hz}. Concurrently, the PD controllers functioned at a higher frequency of 200 \si{Hz}.
During the training process, we set the number of robots (represented as $N$) to 6144. Specifically, as shown in Section \ref{Spatial-DR}, during the initialization of the environment, we first select $\xi_{i,i=0,1,2,3}$ within the range $\mathcal U(0.6,1.4)$. After obtaining $N$ sets of $[\xi_0,\xi_1,\xi_2,\xi_3]$ parameters, we update the original URDF files based on Tab. \ref{URDF} to obtain $N$ sets of different robots, which are then loaded into the simulator.

\begin{table}[t]
	\centering
	\caption{Comparison of domain randomization methods}
	\begin{tabular}{crrrrrr}
		\hline
		Methods & \multicolumn{2}{c}{Temporal-DR} & \multicolumn{2}{c}{Spatial-DR} & \multicolumn{2}{c}{No-DR} \\
		\hline
		Mean value & \multicolumn{2}{c}{3.51$\pm$0.30} & \multicolumn{2}{c}{15.58$\pm$0.67} & \multicolumn{2}{c}{9.91$\pm$0.37} \\
		P-value & \multicolumn{6}{c}{$4.42\times 10^{-5} ~~~~~ 4.55\times 10^{-4} $} \\
		\hline
	\end{tabular}%
	\label{domain}
	\vspace{-2.0em}
\end{table}%

\subsection{Performane of Control Policy}
\textbf{Comparison of Domain Randomization:}
We employed the following three methods to demonstrate the effectiveness of spatial domain randomization (DR).
\textbf{Temporal-DR.} This method varied the leg length of the robots during the training process. Specifically, it uniformly sampled 81 morphologies from the sampling range (as shown in Fig. \ref{heatmap}). It started the training with the first set of morphologies, then switched to the next set once 1/81 of the overall process was completed, and continued this pattern until the training concluded. 
\textbf{Spatial-DR (Ours).} The proposed method, involved using $N$ robots with varied morphologies to simultaneously gather data for training the policy.
\textbf{No-DR.} No domain randomization strategy was employed, and all robots shared the same default morphology, \emph{i.e.}, $\xi_i=[1.0] \times 4$.
We randomly sampled one hundred morphologies and recorded the average cumulative rewards under these morphologies. Furthermore, we conducted T-tests comparing the proposed method with Temporal-DR and No-DR, and the results are presented in Tab. \ref{domain}. It is evident that the proposed method yielded the highest mean values, and the P-values between the proposed method and the two baselines are less than the threshold of 0.05, confirming the statistical superiority over the two baselines.
In addition, we measured the time required by each method. The time taken for Spatial-DR is approximately the same as that for No-DR. In contrast, Temporal-DR is slower, as it requires continuous termination of the current environment and the loading of new morphologies throughout the training process. This procedure is time-consuming, resulting in a training duration that is roughly twice that of the other two methods. These findings indicate that the Spatial-DR method effectively enhances the generalizability of the control policy without requiring additional computational resources.

\begin{figure}[t]
	\centering
	\includegraphics[width=0.45\textwidth]{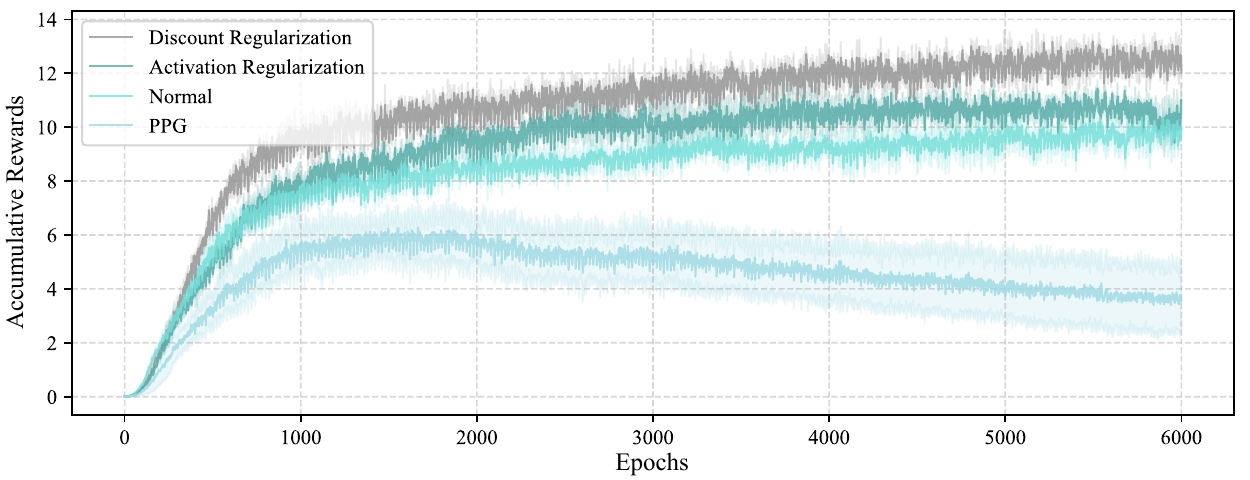}
	\setlength{\abovecaptionskip}{-0.5em}
	\caption{The curve of cumulative rewards changes with the training steps. 
	}
	\vspace{-2.0em}
	\label{reg_res}
\end{figure}

\textbf{Comparison of Regularization (Reg):}
To verify the effectiveness of regularization methods, we designed the following experiments.
\textbf{Activation Reg.} This method added a regularization term to the Critic's loss function. 
\textbf{Discount Reg.} The proposed method, it altered the discount factor $\gamma$ in the PPO algorithm to a smaller value $\gamma_{\rm reg}$. 
\textbf{Normal.} This approach does not employ regularization.
\textbf{Phasic Policy Gradient (PPG) \cite{cobbe2021phasic}.} A SOTA method for enhancing the generalizability of RL algorithms. It integrates auxiliary losses during the training of the PPO algorithm to improve the algorithm's generalizability, details can be found in \cite{cobbe2021phasic}.
We plotted the cumulative reward curves for each approach during the training process, as shown in Fig. \ref{reg_res}. It is evident that D-Reg yielded the best results, followed by A-Reg, while PPG underperformed, even falling behind Normal, which doesn’t use any regularization techniques. 
Taking both the effectiveness and the simplicity of algorithm implementation into account, D-Reg proves to be superior in our locomotion control tasks. It is noteworthy that PPG exhibited the poorest performance. We suspect this may be because the PPG algorithm was developed for image-based RL and might not be well suited for motion control RL with low dimensional observations.

\begin{figure}[t]
	\centering
	\includegraphics[width=0.47\textwidth]{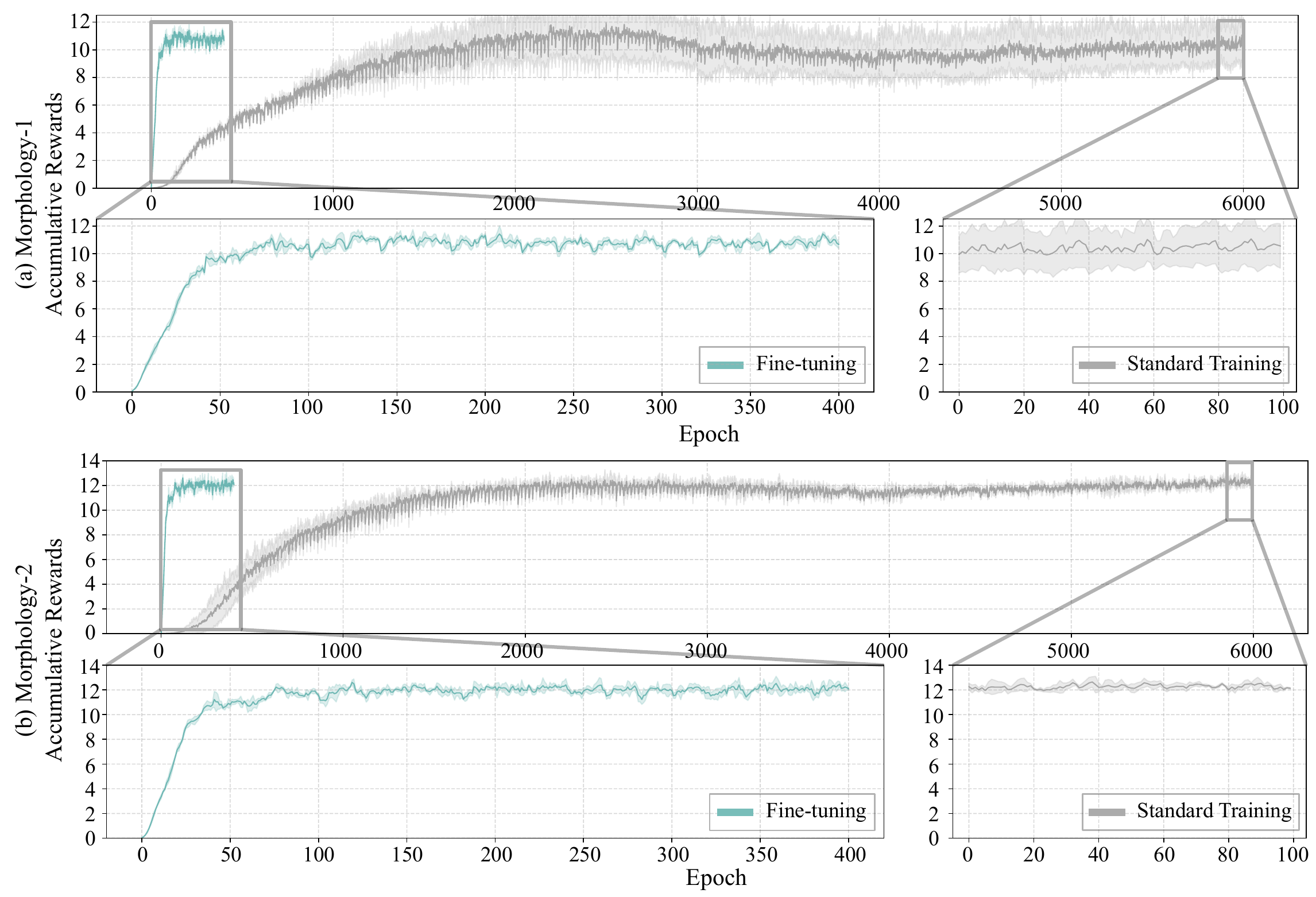}
	\setlength{\abovecaptionskip}{-0.5em}
	\caption{Comparison of Fine-tuning and Standard Training. (a) and (b) represent the results under different morphologies.}
	\vspace{-2.0em}
	\label{fine_com}
\end{figure}

\textbf{Performance of Fine-tuning:}
In this section, we demonstrate the effectiveness of fine-tuning, we applied two methods as follows.
\textbf{Standard Training.} Normative training under the specific morphology without pre-training.
\textbf{Fine-tuning.} The proposed method.
The training time comparison, illustrated in Fig. \ref{fine_com}, reveals that Standard Training begins to converge at the 2000th epoch, while Fine-tuning starts converging at just the 100th epoch. These findings validate that the proposed method can significantly accelerate the training process. For instance, when training 30 specific morphologies, Standard Training requires $30 \times 6000 =1.8\times10^{5}$ epochs, whereas Fine-tuning only needs $6000 + 30 \times 400 = 1.8\times 10^4$ epochs. The number 6000 represents the pre-training steps, while 400 indicates the fine-tuning steps tailored for specific morphologies, equivalent to 6.67\% of the former.
Furthermore, we conduct a performance analysis of the two methods, as illustrated in Fig. \ref{fine_com}. For Morphology-1, both methods exhibit convergence around a value of 10, while for Morphology-2, convergence is observed around a value of 12. We calculate the mean of the last 100 reward values for both methods and assess the differences. Specifically, for Morphology-1, the difference in reward values between the two methods is 2.88\%, whereas for Morphology-2, this difference is 2.08\%. These findings indicate that the proposed approach can achieve performance comparable to standard training within a smaller margin of error, while requiring only one-tenth of the time.

\subsection{Comparison with Co-design Baselines}
In this section, we compare the proposed method with three co-design baselines, we first introduce these methods.
\textbf{Online-PCODP \cite{bjelonic2023learning}.} This method treats design parameters as privileged information and samples these parameters uniformly within their range during the training process. It assumes that near-optimal performance can be achieved using a policy conditioned on design parameters. 
\textbf{Offline-PCODP \cite{chen2023c}.} 
This method collects interaction data from various morphologies using online RL and then uses offline RL to create a general policy from these data. 
We employed 16 distinct morphologies to gather interaction data, utilizing the TD3-BC \cite{fujimoto2021minimalist} algorithm for offline training.
\textbf{EAT \cite{yu2023multi}.} 
Similar to Offline-PCODP, the primary difference lies in replacing the TD3-BC method with the Decision Transformer \cite{chen2021decision}.
\textbf{Ours.} The proposed method.

\begin{figure*}[t]
	\centering
	\includegraphics[width=1.0\textwidth]{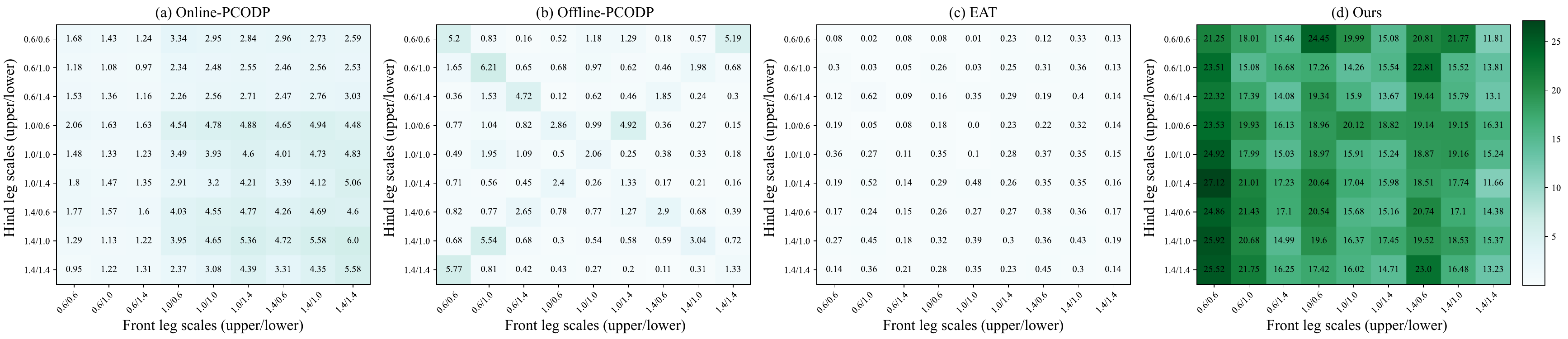}
	\setlength{\abovecaptionskip}{-1.8em}
	\caption{Heatmap of co-design experiments. Each block represents the cumulative rewards achieved using the corresponding control strategy under a given morphology.
	For Offline-PCOD and EAT, the 16 typical morphologies are located along the diagonal of the heatmap (excluding the center point).
	}
	\vspace{-1.0em}
	\label{heatmap}
\end{figure*}

We uniformly sampled 81 different morphologies and plotted the cumulative rewards of different methods in Fig. \ref{heatmap}.
As can be seen from Fig. \ref{heatmap}(d), our proposed method achieves the highest rewards across various morphologies. For the Online-PCODP method, as observed in Fig. \ref{heatmap}(a), its highest reward values are concentrated under the last trained morphology. For previously encountered morphologies, it shows lower cumulative rewards. For the Offline-PCODP method, as seen in Fig. \ref{heatmap}(b), its highest rewards are found near the 16 typical morphologies encountered during training. However, for morphologies not previously encountered, it shows lower cumulative rewards. Regarding the EAT method, as seen in Fig. \ref{heatmap}(c), its performance appears to be poor across all morphologies. We speculate that this may be due to the original EAT method focusing on simple planar walking tasks with low-dimensional states such as body linear velocity and angular velocity. In the parkour task, terrain information is crucial but results in higher state dimensionality, potentially causing Transformer-based algorithms to fail.

\begin{figure*}[t]
	\centering
	\includegraphics[width=1.0\textwidth]{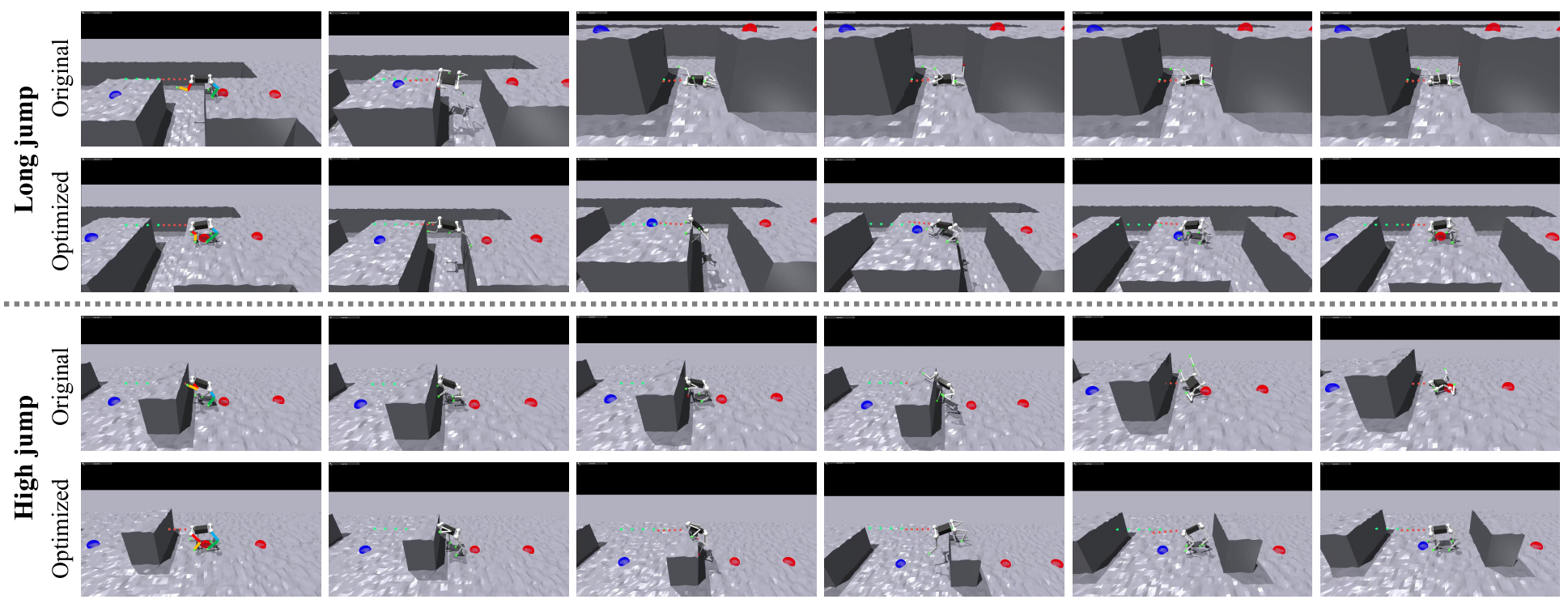}
	\setlength{\abovecaptionskip}{-1.5em}
	\caption{Comparison screenshots of different morphologies in extreme long jump and high jump tasks. For each task, the first row displays the original morphology, while the second row shows the optimized morphology. The sampling interval is 0.5\si{s}.}
	\vspace{-1.5em}
	\label{snapshot}
\end{figure*}

Furthermore, we applied the aforementioned methods for co-design in both the high jump and long jump tasks. We recorded the accumulative rewards in Tab. \ref{baselines}. It is evident that the proposed method achieved the highest values in both tasks, markedly surpassing the baseline methods.
In Tab. \ref{baselines}, we also recorded the time consumption of various methods. The introduction of the fine-tuning during the BO process has incurred some additional time compared to Online-PCODP method, however, overall, the time consumed by the proposed method is still much less than that of the Offline-PCODP and EAT methods. Taking both performance and time consumption into consideration, the advantages of the proposed method are clearly notable.

\begin{table}[t]
	\centering
	\caption{Results of comparison with co-design baselines}
	\vspace{-1em}
	\resizebox{0.40\textwidth}{!}{
		\begin{tabular}{ccc|c}
			\hline
			\diagbox{Method}{Task}& Long jump & High jump & \makecell[c]{Time \\ Consumption} \\
			\hline
			Online-PCODP \cite{bjelonic2023learning} &   \underline{6.18}    &  \underline{10.37}    & \bf{3\si{h} 5\si{min}} \\
			Offline-PCODP \cite{chen2023c} &   4.78    &  4.63     & 86\si{h} 45\si{min}  \\
			EAT  \cite{yu2023multi} &   0.66    &    0.71   &  86\si{h} 17\si{min} \\
			Ours  &   \textbf{28.23}   &   \textbf{29.06}    & \underline{8\si{h} 51\si{min}}  \\
			\hline
		\end{tabular}%
		\label{baselines}}
	\vspace{-2.5em}
\end{table}%

\subsection{Parkour Performance Comparison}
In this section, we compared the performance achievable by baselines with that of the proposed approach for the tasks of long jump and high jump, as illustrated in Tab. \ref{baselines-2}. The values for the baseline methods were cited from the corresponding papers. As seen in Tab. \ref{baselines-2}, the proposed method significantly enhances the performance of robots in various parkour tasks, demonstrating that, in addition to improving the motion controller's performance, appropriate optimization of the mechanical structure can also enhance the performance of quadruped robots in extreme tasks. Furthermore, we illustrated the performance of the robots in extreme tasks before and after morphological optimization, as shown in Fig. \ref{snapshot}. For the long jump task, it is clear that the optimized robots are equipped with longer hind legs, which can provide more robust propulsion, while shorter front legs can assist the robot in gripping the ground ahead to prevent falls. Similarly, for the high jump task, the robots evolved to have longer legs, aiding in climbing. Both experiments underscore the potential of the co-design approach in unlocking the parkour performance of robots.

\begin{table}[t]
	\centering
	\caption{Parkour performance comparison}
	\vspace{-1em}
	\resizebox{0.40\textwidth}{!}{
		\begin{tabular}{ccc}
			\hline
			\diagbox{Method}{Task}& Long jump & High jump \\
			\hline
			Robot Parkour Learning\cite{zhuang2023robot} &   0.6\si{m}   & 0.4\si{m}  \\
			Extreme Parkour\cite{cheng2023extreme} &    0.8\si{m}   & 0.5\si{m} \\
			Ours  &   \bf{1.0\si{m}}    & \bf{0.55\si{m}} \\
			\hline
		\end{tabular}
		\label{baselines-2}}
	\vspace{-2.5em}
\end{table}%

\section{Conclusions}
In this paper, we introduced a pretraining-finetuning framework for mechanical-control co-design in robotics. During the pre-training phase, we proposed a combination of spatial domain randomization and regularization method to train a generalizable model. We embedded the fine-tuning stage within the iterative process of BO, allowing for rapid, morphology-specific learning based on the pre-trained model during each round of BO. This approach ensures the optimality of policies across different morphologies while being time-efficient. By integrating mechanical structure optimization with control policies, we have taken a significant step forward in enhancing the extreme motion capabilities of quadruped robots, opening new avenues for their application and effectiveness in diverse environments. 
Currently, due to limitations in hardware structure, we have not yet created a physical robot with adjustable leg lengths. In future work, we will address this issue and validate the proposed algorithm in real-world applications.
\bibliographystyle{IEEEtran}
\bibliography{ref}

\begin{thebibliography}{10}
\providecommand{\url}[1]{#1}
\csname url@samestyle\endcsname
\providecommand{\newblock}{\relax}
\providecommand{\bibinfo}[2]{#2}
\providecommand{\BIBentrySTDinterwordspacing}{\spaceskip=0pt\relax}
\providecommand{\BIBentryALTinterwordstretchfactor}{4}
\providecommand{\BIBentryALTinterwordspacing}{\spaceskip=\fontdimen2\font plus
\BIBentryALTinterwordstretchfactor\fontdimen3\font minus
  \fontdimen4\font\relax}
\providecommand{\BIBforeignlanguage}[2]{{%
\expandafter\ifx\csname l@#1\endcsname\relax
\typeout{** WARNING: IEEEtran.bst: No hyphenation pattern has been}%
\typeout{** loaded for the language `#1'. Using the pattern for}%
\typeout{** the default language instead.}%
\else
\language=\csname l@#1\endcsname
\fi
#2}}
\providecommand{\BIBdecl}{\relax}
\BIBdecl

\bibitem{gupta2021embodied}
A.~Gupta, S.~Savarese, S.~Ganguli, and L.~Fei-Fei, ``Embodied intelligence via
  learning and evolution,'' \emph{Nature communications}, vol.~12, no.~1, p.
  5721, 2021.

\bibitem{belmonte2022meta}
{\'A}.~Belmonte-Baeza, J.~Lee, G.~Valsecchi, and M.~Hutter, ``Meta
  reinforcement learning for optimal design of legged robots,'' \emph{IEEE
  Robotics and Automation Letters}, vol.~7, no.~4, pp. 12\,134--12\,141, 2022.

\bibitem{chen2023c}
C.~Chen, P.~Xiang, H.~Lu, Y.~Wang, and R.~Xiong, ``C 2: Co-design of robots via
  concurrent-network coupling online and offline reinforcement learning,'' in
  \emph{2023 IEEE/RSJ International Conference on Intelligent Robots and
  Systems (IROS)}.\hskip 1em plus 0.5em minus 0.4em\relax IEEE, 2023, pp.
  7487--7494.

\bibitem{chen2023deep}
C.~Chen, P.~Xiang, J.~Zhang, R.~Xiong, Y.~Wang, and H.~Lu, ``Deep reinforcement
  learning based co-optimization of morphology and gait for small-scale legged
  robot,'' \emph{IEEE/ASME Transactions on Mechatronics}, 2023.

\bibitem{yu2023multi}
C.~Yu, W.~Zhang, H.~Lai, Z.~Tian, L.~Kneip, and J.~Wang, ``Multi-embodiment
  legged robot control as a sequence modeling problem,'' in \emph{2023 IEEE
  International Conference on Robotics and Automation (ICRA)}.\hskip 1em plus
  0.5em minus 0.4em\relax IEEE, 2023, pp. 7250--7257.

\bibitem{bjelonic2023learning}
F.~Bjelonic, J.~Lee, P.~Arm, D.~Sako, D.~Tateo, J.~Peters, and M.~Hutter,
  ``Learning-based design and control for quadrupedal robots with
  parallel-elastic actuators,'' \emph{IEEE Robotics and Automation Letters},
  vol.~8, no.~3, pp. 1611--1618, 2023.

\bibitem{zhuang2023robot}
Z.~Zhuang, Z.~Fu, J.~Wang, C.~G. Atkeson, S.~Schwertfeger, C.~Finn, and
  H.~Zhao, ``Robot parkour learning,'' in \emph{Conference on Robot
  Learning}.\hskip 1em plus 0.5em minus 0.4em\relax PMLR, 2023, pp. 73--92.

\bibitem{hoeller2023anymal}
D.~Hoeller, N.~Rudin, D.~Sako, and M.~Hutter, ``Anymal parkour: Learning agile
  navigation for quadrupedal robots,'' \emph{arXiv preprint arXiv:2306.14874},
  2023.

\bibitem{cheng2023extreme}
X.~Cheng, K.~Shi, A.~Agarwal, and D.~Pathak, ``Extreme parkour with legged
  robots,'' in \emph{RoboLetics: Workshop on Robot Learning in Athletics@ CoRL
  2023}, 2023.

\bibitem{rudin2022advanced}
N.~Rudin, D.~Hoeller, M.~Bjelonic, and M.~Hutter, ``Advanced skills by learning
  locomotion and local navigation end-to-end,'' in \emph{2022 IEEE/RSJ
  International Conference on Intelligent Robots and Systems (IROS)}.\hskip 1em
  plus 0.5em minus 0.4em\relax IEEE, 2022, pp. 2497--2503.

\bibitem{ha2017joint}
S.~Ha, S.~Coros, A.~Alspach, J.~Kim, and K.~Yamane, ``Joint optimization of
  robot design and motion parameters using the implicit function theorem.'' in
  \emph{Robotics: Science and systems}, vol.~8, 2017.

\bibitem{geilinger2018skaterbots}
M.~Geilinger, R.~Poranne, R.~Desai, B.~Thomaszewski, and S.~Coros,
  ``Skaterbots: Optimization-based design and motion synthesis for robotic
  creatures with legs and wheels,'' \emph{ACM Transactions on Graphics (TOG)},
  vol.~37, no.~4, pp. 1--12, 2018.

\bibitem{fadini2021computational}
G.~Fadini, T.~Flayols, A.~Del~Prete, N.~Mansard, and P.~Sou{\`e}res,
  ``Computational design of energy-efficient legged robots: Optimizing for size
  and actuators,'' in \emph{2021 IEEE International Conference on Robotics and
  Automation (ICRA)}.\hskip 1em plus 0.5em minus 0.4em\relax IEEE, 2021, pp.
  9898--9904.

\bibitem{hejna2020task}
D.~J. Hejna~III, P.~Abbeel, and L.~Pinto, ``Task-agnostic morphology
  evolution,'' in \emph{International Conference on Learning Representations},
  2020.

\bibitem{wang2018neural}
T.~Wang, Y.~Zhou, S.~Fidler, and J.~Ba, ``Neural graph evolution: Towards
  efficient automatic robot design,'' in \emph{International Conference on
  Learning Representations}, 2018.

\bibitem{luck2020data}
K.~S. Luck, H.~B. Amor, and R.~Calandra, ``Data-efficient co-adaptation of
  morphology and behaviour with deep reinforcement learning,'' in
  \emph{Conference on Robot Learning}.\hskip 1em plus 0.5em minus 0.4em\relax
  PMLR, 2020, pp. 854--869.

\bibitem{schaff2019jointly}
C.~Schaff, D.~Yunis, A.~Chakrabarti, and M.~R. Walter, ``Jointly learning to
  construct and control agents using deep reinforcement learning,'' in
  \emph{2019 International Conference on Robotics and Automation (ICRA)}.\hskip
  1em plus 0.5em minus 0.4em\relax IEEE, 2019, pp. 9798--9805.

\bibitem{nagarajan2009state}
U.~Nagarajan, A.~Mampetta, G.~A. Kantor, and R.~L. Hollis, ``State transition,
  balancing, station keeping, and yaw control for a dynamically stable single
  spherical wheel mobile robot,'' in \emph{2009 IEEE International Conference
  on Robotics and Automation}.\hskip 1em plus 0.5em minus 0.4em\relax IEEE,
  2009, pp. 998--1003.

\bibitem{rudin2022learning}
N.~Rudin, D.~Hoeller, P.~Reist, and M.~Hutter, ``Learning to walk in minutes
  using massively parallel deep reinforcement learning,'' in \emph{Conference
  on Robot Learning}.\hskip 1em plus 0.5em minus 0.4em\relax PMLR, 2022, pp.
  91--100.

\bibitem{kumar2021rma}
A.~Kumar, Z.~Fu, D.~Pathak, and J.~Malik, ``Rma: Rapid motor adaptation for
  legged robots,'' \emph{arXiv preprint arXiv:2107.04034}, 2021.

\bibitem{ji2022concurrent}
G.~Ji, J.~Mun, H.~Kim, and J.~Hwangbo, ``Concurrent training of a control
  policy and a state estimator for dynamic and robust legged locomotion,''
  \emph{IEEE Robotics and Automation Letters}, vol.~7, no.~2, pp. 4630--4637,
  2022.

\bibitem{nahrendra2023dreamwaq}
I.~M.~A. Nahrendra, B.~Yu, and H.~Myung, ``Dreamwaq: Learning robust
  quadrupedal locomotion with implicit terrain imagination via deep
  reinforcement learning,'' in \emph{2023 IEEE International Conference on
  Robotics and Automation (ICRA)}.\hskip 1em plus 0.5em minus 0.4em\relax IEEE,
  2023, pp. 5078--5084.

\bibitem{shi2022reinforcement}
H.~Shi, B.~Zhou, H.~Zeng, F.~Wang, Y.~Dong, J.~Li, K.~Wang, H.~Tian, and
  M.~Q.-H. Meng, ``Reinforcement learning with evolutionary trajectory
  generator: A general approach for quadrupedal locomotion,'' \emph{IEEE
  Robotics and Automation Letters}, vol.~7, no.~2, pp. 3085--3092, 2022.

\bibitem{tan2018sim}
J.~Tan, T.~Zhang, E.~Coumans, A.~Iscen, Y.~Bai, D.~Hafner, S.~Bohez, and
  V.~Vanhoucke, ``Sim-to-real: Learning agile locomotion for quadruped
  robots,'' \emph{arXiv preprint arXiv:1804.10332}, 2018.

\bibitem{makoviychuk2021isaac}
V.~Makoviychuk, L.~Wawrzyniak, Y.~Guo, M.~Lu, K.~Storey, M.~Macklin,
  D.~Hoeller, N.~Rudin, A.~Allshire, A.~Handa \emph{et~al.}, ``Isaac gym: High
  performance gpu based physics simulation for robot learning,'' in
  \emph{Thirty-fifth Conference on Neural Information Processing Systems
  Datasets and Benchmarks Track (Round 2)}, 2021.

\bibitem{luo2024moral}
Z.~Luo, Y.~Dong, X.~Li, R.~Huang, Z.~Shu, E.~Xiao, and P.~Lu, ``Moral: Learning
  morphologically adaptive locomotion controller for quadrupedal robots on
  challenging terrains,'' \emph{IEEE Robotics and Automation Letters}, 2024.

\bibitem{moon2022rethinking}
S.~Moon, J.~Lee, and H.~O. Song, ``Rethinking value function learning for
  generalization in reinforcement learning,'' \emph{Advances in Neural
  Information Processing Systems}, vol.~35, pp. 34\,846--34\,858, 2022.

\bibitem{cobbe2021phasic}
K.~W. Cobbe, J.~Hilton, O.~Klimov, and J.~Schulman, ``Phasic policy gradient,''
  in \emph{International Conference on Machine Learning}.\hskip 1em plus 0.5em
  minus 0.4em\relax PMLR, 2021, pp. 2020--2027.

\bibitem{schulman2015high}
J.~Schulman, P.~Moritz, S.~Levine, M.~Jordan, and P.~Abbeel, ``High-dimensional
  continuous control using generalized advantage estimation,'' \emph{arXiv
  preprint arXiv:1506.02438}, 2015.

\bibitem{fujimoto2021minimalist}
S.~Fujimoto and S.~S. Gu, ``A minimalist approach to offline reinforcement
  learning,'' \emph{Advances in neural information processing systems},
  vol.~34, pp. 20\,132--20\,145, 2021.

\bibitem{chen2021decision}
L.~Chen, K.~Lu, A.~Rajeswaran, K.~Lee, A.~Grover, M.~Laskin, P.~Abbeel,
  A.~Srinivas, and I.~Mordatch, ``Decision transformer: Reinforcement learning
  via sequence modeling,'' \emph{Advances in neural information processing
  systems}, vol.~34, pp. 15\,084--15\,097, 2021.

\end{thebibliography}

\clearpage
\appendix
\subsection{Reward Function}
\label{reward_function}

\begin{table}[!htbp]
	\centering
	\caption{Reward function elements}
	\begin{threeparttable}
		\resizebox{0.48\textwidth}{!}{
			\begin{tabular}{cccc}
			\hline
			Reward & Equation  & Weight  \\ 
			\hline
			Goal Tracking  & $\begin{array}{l}
			\min \left( {\left\langle {{\mathbf v},\mathbf{{\hat d}_w}} \right\rangle ,v^{cmd}} \right)\\
			{\mathbf{{\hat d}_w}} = \frac{{\mathbf p - \mathbf x}}{{\left\| {\mathbf p - \mathbf x} \right\|}}
			\end{array}$  & 1.5   \\
			Clearance & $ - \sum {_{i = 0}^4{c_i} \cdot M\left[ {{p_i}} \right]} $ & -1.0 \\
			Yaw Tracking & $\exp \left\{ { - {{\left\| {{\omega _z} - \omega _z^{des}} \right\|}_1}} \right\}$    & 0.5  \\
			Linear Velocity ($z$)  & $ v_z^2$    & -1.0   \\
			Angular Velocity ($xy$) & $\omega _{xy}^2$ & -0.05 \\
			Action Rate & ${\left\| {{a_t} - {a_{t - 1}}} \right\|_2}$ & -0.1 \\
			Hip Position & ${\left( {{q_{hip}} - q_{hip}^{des}} \right)^2}$ & -0.5 \\
			
			Joint Acceleration & ${\ddot q^2}$ & $-2.5\times 10^{-7}$ \\
			Joint Cosmetic & ${\left( {q - {q^{des}}} \right)^2}$ & -0.04 \\
			Torque Change & ${\left( {{\tau _t} - {\tau _{t - 1}}} \right)^2}$ & $-1\times10^{-7}$ \\
			Torque Penalty & $\tau _t^2$ & $-1\times10^{-5}$ \\
			Orientation & $g^{2}$ & -1.0 \\
			\hline
			\end{tabular}
		}
		\begin{tablenotes}
			\footnotesize
			\begin{tabular}{p{0.48\textwidth}} 
				\item In this description, $\mathbf{p}$ represents the subsequent waypoint, while $\mathbf{x}$ denotes the robot's location within the world frame. The robot's current velocity in the world frame is expressed as $\mathbf v \in \mathbb{R}^2$, and the target speed is indicated by ${v^{cmd}} \in \mathbb{R}^1$. The term $c_i$ equals 1 when the $i$-th foot is in contact with the ground. The boolean function $M$ returns 1 if the point $p_i$ is situated within 5\si{cm} of an edge. Each leg's foot position is given by $p_i$. The notation $\exp(\cdot)$ refers to exponential operators, and the superscript ${(\cdot)^{des}}$ is used to specify desired values. The base angular velocity is represented as $\omega \in \mathbb{R}^3$, and the base linear velocity as $v \in \mathbb{R}^3$. The actions are denoted by ${a_t} \in \mathbb{R}^{12}$, the joint angles by $q \in \mathbb{R}^{12}$, the joint accelerations by $\ddot{q} \in \mathbb{R}^{12}$, and the joint torques by ${\tau} \in \mathbb{R}^{12}$. Lastly, $g$ represents the gravity vector projected into the robot's body frame.
			\end{tabular}
		\end{tablenotes}  
	\end{threeparttable}
	\label{rewards}
	\vspace{-2em}
\end{table}

\subsection{Regularization}
\label{regularization}
To illustrate, we provide an example with $\lambda=0$, where the loss function for the critic is shown as follows:
\begin{equation}
\resizebox{1\hsize}{!}{$
\begin{array}{l}
{J_V}(\phi ) = {\mathbb E_{{s_t} \sim \tau }}\left[ {\frac{1}{2}{{\left( {{V_\phi }({s_t}) - {{\hat R}_t}} \right)}^2}} \right]\\
~~~~~~~~= {\mathbb E_{{s_t} \sim \tau }}\left[ {\frac{1}{2}{{\left( {{V_\phi }({s_t}) - ({{\hat A}_t} + {V_{{\phi _{old}}}}({s_t}))} \right)}^2}} \right]\\
~~~~~~~~\overset{\textcircled{1}}{\mathop{=}} {\mathbb E_{{s_t} \sim \tau }}\left[ {\frac{1}{2}{{\left( {{V_\phi }({s_t}) - \left( {{r_t} + \gamma {V_\phi }({s_{t + 1}})} \right)} \right)}^2}} \right]
\end{array} $} \label{eq_5}
\end{equation}
\textcircled{1} is valid exclusively when $\lambda = 0$. According to Eq. (\ref{eq_5}), the update formula for $\phi_i$ is as follows:
\begin{equation}
\resizebox{1\hsize}{!}{$
{\phi _{i + 1}} = {\phi _i} + {\alpha _i}({r_t} + \gamma {V_{{\phi _i}}}({s_{t + 1}}) - {V_{{\phi _i}}}({s_t}))\nabla {V_{{\phi _i}}}({s_t}) $}
\end{equation}
where $\alpha_i$ is the learning rate. Consider $\omega$ as the parameter sequence generated by the PPO algorithm using the discounted factor $\gamma_{\rm{reg}}$, together with the inclusion of the regularization function $\beta(\hat V_{\phi}(s))$, where $\beta$ is given by $\frac{{{\gamma _{\rm{reg}}} - \gamma }}{{2\gamma }}$. Additionally, this setup involves reward scaling by a factor of $\frac{{{\gamma _{\rm{reg}}}}}{\gamma }$, an adjusted learning rate of ${\alpha '_i} = \frac{\gamma }{{{\gamma _{\rm{reg}}}}}{\alpha _i}$, and the same initial parameters $\phi_0$. We aim to demonstrate by induction that ${\phi _i} = {\omega _i},i = 1,2,...$.

The base case, where $\omega_{0}=\phi_{0}$, is directly established by equal initial settings. Assuming that $\omega_i=\phi_i$. We proceed to prove the induction step for $i+1$. The update formula for the $i$-th step can be reformulated as follows:
\begin{align}
{\phi _{i + 1}} =& {\phi _i} + {\alpha _i}({r_t} + \gamma {V_{{\phi _i}}}({s_{t + 1}}) - {V_{{\phi _i}}}({s_t}))\nabla {V_{{\phi _i}}}({s_t})\nonumber\\
\overset{\textcircled{2}}{\mathop{=}}\,&{\omega _i} + {\alpha _i}({r_t} + \gamma {V_{{\omega _i}}}({s_{t + 1}}) - {V_{{\omega _i}}}({s_t}))\nabla {V_{{\omega _i}}}({s_t}) \nonumber\\
=& {\omega _i} + {\alpha _i}\frac{\gamma }{{{\gamma _{{\rm{reg}}}}}}\left( \frac{{{\gamma _{{\rm{reg}}}}}}{\gamma }{r_t} + {\gamma _{{\rm{reg}}}}{V_{{\omega _i}}}({s_{t + 1}})\right.\nonumber\\ 
&\quad\qquad\left.- \frac{{{\gamma _{re}}}}{\gamma }{V_{{\omega _i}}}({s_t}) \right)\nabla {V_{{\omega _i}}}({s_t}) \nonumber\\
{\mathop{=}}\,&{\omega _i} + {{\alpha '}_i}\left( \frac{{{\gamma _{{\rm{reg}}}}}}{\gamma }{r_t} + {\gamma _{{\rm{reg}}}}{V_{{\omega _i}}}({s_{t + 1}}) - {V_{{\omega _i}}}({s_t}) \right. \nonumber\\
&\quad\qquad\left.	+ {V_{{\omega _i}}}({s_t}) -  \frac{{{\gamma _{{\rm{reg}}}}}}{\gamma }{V_{{\omega _i}}}({s_t}) \right)\nabla {V_{{\omega _i}}}({s_t}) \nonumber\\
=& {\omega _i} + {{\alpha '}_i}\left( {\frac{{{\gamma _{{\rm{reg}}}}}}{\gamma }{r_t} + {\gamma _{{\rm{reg}}}}{V_{{\omega _i}}}({s_{t + 1}}) - {V_{{\omega _i}}}({s_t})} \right)\nabla {V_{{\omega _i}}}({s_t}) \nonumber\\
&- {{\alpha '}_i}\frac{{{\gamma _{{\rm{reg}}}} - \gamma }}{\gamma }{V_{{\omega _i}}}({s_t})\nabla {V_{{\omega _i}}}({s_t}) \nonumber\\
=& {\omega _i} + {{\alpha '}_i}\left( {\frac{{{\gamma _{{\rm{reg}}}}}}{\gamma }{r_t} + {\gamma _{{\rm{reg}}}}{V_{{\omega _i}}}({s_{t + 1}}) - {V_{{\omega _i}}}({s_t})} \right)\nabla {V_{{\omega _i}}}({s_t}) \nonumber\\
&- {{\alpha '}_i}\nabla \left( {\frac{{{\gamma _{{\rm{reg}}}} - \gamma }}{{2\gamma }}{{\left( {{V_{{\omega _i}}}({s_t})} \right)}^2}} \right) \nonumber\\
=& {\omega _{i + 1}} 
\end{align}
where \textcircled{2} is workable due to the induction assumption.

\end{document}